\documentclass[letterpaper, 10 pt, conference]{ieeeconf}

\usepackage{cite}
\usepackage{microtype}
\usepackage{amsmath,amssymb,amsfonts}
\usepackage{algorithmic}
\usepackage{graphicx}
\usepackage{textcomp}
\usepackage{xcolor}
\usepackage[dvipsnames]{xcolor}
\usepackage{todonotes}
\usepackage{tikz}
\usepackage{subcaption}
\usepackage{lipsum}
\usepackage{xparse} 
\usepackage{svg}      
\usepackage{float} 
\usepackage{colortbl}
\usepackage[table]{xcolor}
\usepackage{dsfont}
\usepackage{bm}
\usepackage{siunitx}    

\DeclareMathOperator{\minADE}{minADE}

\usetikzlibrary{arrows.meta, positioning, calc}

\def\BibTeX{{\rm B\kern-.05em{\sc i\kern-.025em b}\kern-.08em
    T\kern-.1667em\lower.7ex\hbox{E}\kern-.125emX}}
\begin{document}

\title{
\LARGE \bf
Super Agents and Confounders: Influence of surrounding agents on vehicle trajectory prediction\\
}

\newif\ifredacted

\ifredacted
    \author{Redacted for review}
\else
    \author{
        Daniel Jost\textsuperscript{1}, 
        Luca Paparusso\textsuperscript{2}, 
        Martin Stoll\textsuperscript{2}, 
        Jörg Wagner\textsuperscript{2}, 
        Raghu Rajan\textsuperscript{1}, 
        Joschka Bödecker\textsuperscript{1}
    }
\fi

\maketitle
\thispagestyle{empty}
\pagestyle{empty}

\ifredacted
\else
    \begingroup
    \renewcommand\thefootnote{\textsuperscript{\arabic{footnote}}}
    \footnotetext[1]{Department of Computer Science, University of Freiburg, Germany. \texttt{\{jostd, rajanr, jboedeck\}@informatik.uni-freiburg.de}}
    \footnotetext[2]{Bosch Center for Artificial Intelligence, Renningen, Germany. \texttt{\{luca.paparusso, martin.stoll, joerg.wagner3\}@de.bosch.com}}
    \endgroup
\fi


\begin{tikzpicture}[remember picture,overlay]
  \node[anchor=north west, xshift=1.6cm, yshift=-0.8cm] at (current page.north west) {
    \small \bfseries \begin{tabular}{l}
    2026 IEEE/RSJ International Conference on Intelligent Robots and Systems (IROS) \\
    September 27-October 1, 2026. Pittsburgh, PA, USA
    \end{tabular}
  };
  \node[anchor=south west, xshift=1.4cm, yshift=1.2cm] at (current page.south west) {
    \footnotesize \textbf{Copyright \copyright 2026 IEEE}
  };
\end{tikzpicture}

\begin{abstract}

In highly interactive driving scenes, trajectory prediction is conditioned on information from surrounding traffic participants such as cars and pedestrians.
Our main contribution is a comprehensive analysis of state-of-the-art trajectory predictors, which reveals a surprising and critical flaw: many surrounding agents degrade prediction accuracy rather than improve it. Using Shapley-based attribution, we rigorously demonstrate that models learn unstable and non-causal decision-making schemes that vary significantly across training runs.
Building on these insights, we propose to integrate a Conditional Information Bottleneck (CIB), which does not require additional supervision and is trained to effectively compress agent features as well as ignore those that are not beneficial for the prediction task.
Comprehensive experiments using multiple datasets and model architectures demonstrate that this simple yet effective approach not only improves overall trajectory prediction performance in many cases but also increases robustness to different perturbations.
Our results highlight the importance of selectively integrating contextual information, which can often contain spurious or misleading signals, in trajectory prediction. Moreover, we provide interpretable metrics for identifying non-robust behavior and present a promising avenue towards a solution.

\end{abstract}

\section{Introduction}
Predicting the future trajectories of agents in dynamic environments is a central task in autonomous driving, crowd simulation, and robotics. Accurate prediction lets autonomous systems anticipate others’ behavior and plan safe, efficient maneuvers. Recent learning-based approaches have achieved strong performance in this domain \cite{chenQEANetImplicitSocial2024, liuLAformerTrajectoryPrediction2023, shiMTRMultiAgentMotion2024, linEDAEvolvingDistinct2023}.

Most trajectory prediction models explicitly incorporate information about surrounding agents, under the assumption that additional context improves accuracy. From a causal perspective, however, not all agents are equally relevant for forecasting a particular target agent’s behavior. A robust model should therefore prioritize relevant agents while discounting irrelevant ones. Yet, recent work shows that removing seemingly irrelevant agents can dramatically change the model’s predictions \cite{roelofsCausalAgentsRobustnessBenchmark2022}, and that small perturbations in agents’ past trajectories can be exploited to increase the likelihood of collisions \cite{zhangAdversarialRobustnessTrajectory2022, saadatnejadAreSociallyawareTrajectory2022, caoAdvDORealisticAdversarial2022}. These findings suggest that state-of-the-art methods are fragile, relying on agent information in a way that is neither selective nor robust.
\begin{figure}[htb]
    \centering
    
    \begin{subfigure}{\linewidth}
        \centering
        \includegraphics[width=0.9\linewidth, trim={0cm 0.5cm 0cm 0cm}, clip]{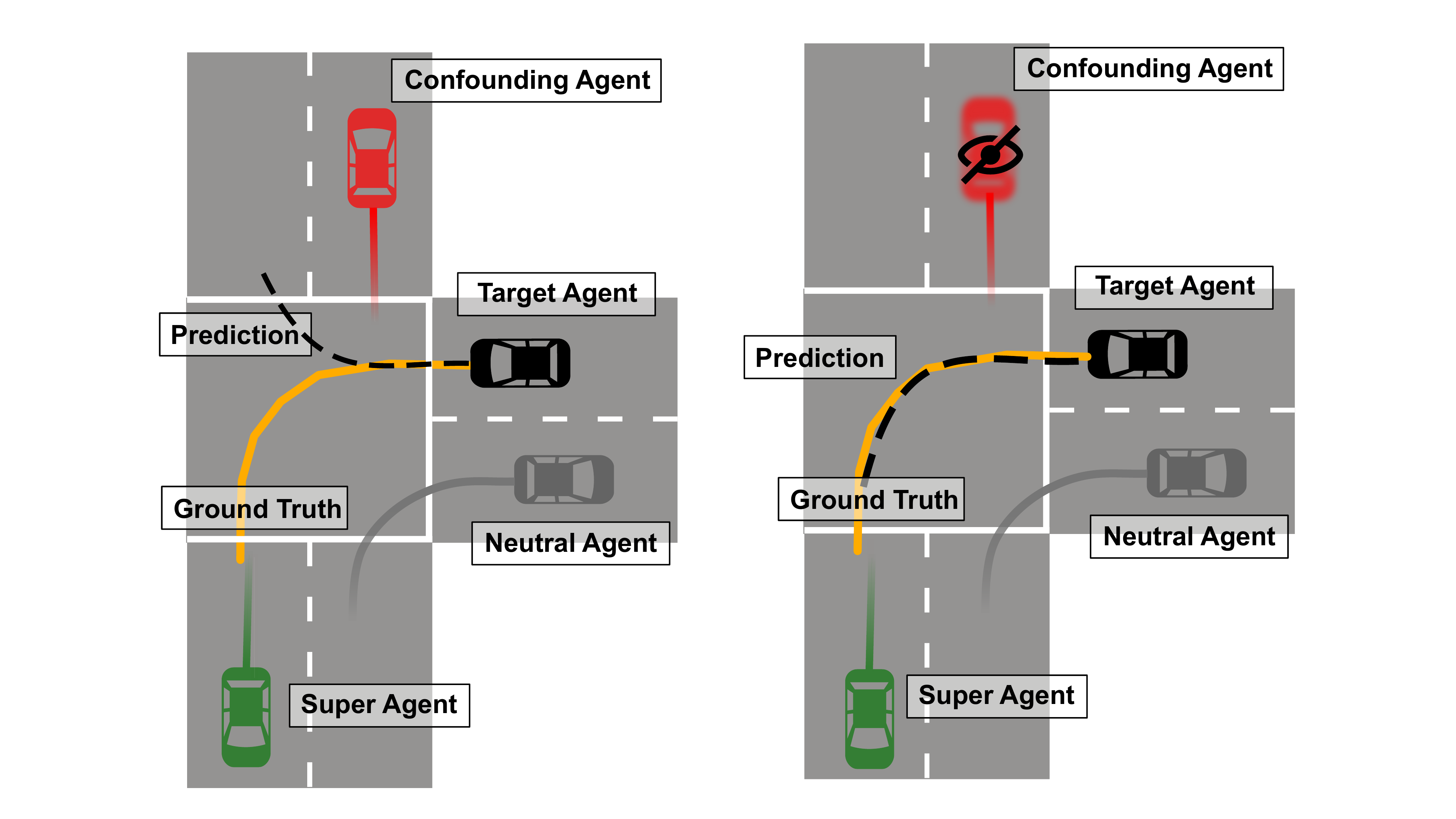}
        \caption{Impact of Confounding Agent removal.}        
        \label{fig:page_one_fig}
    \end{subfigure}
    
    \begin{subfigure}{\linewidth}
        \centering
            \includegraphics[width=0.95\linewidth, trim={0cm 0cm 0cm 0cm}, clip]{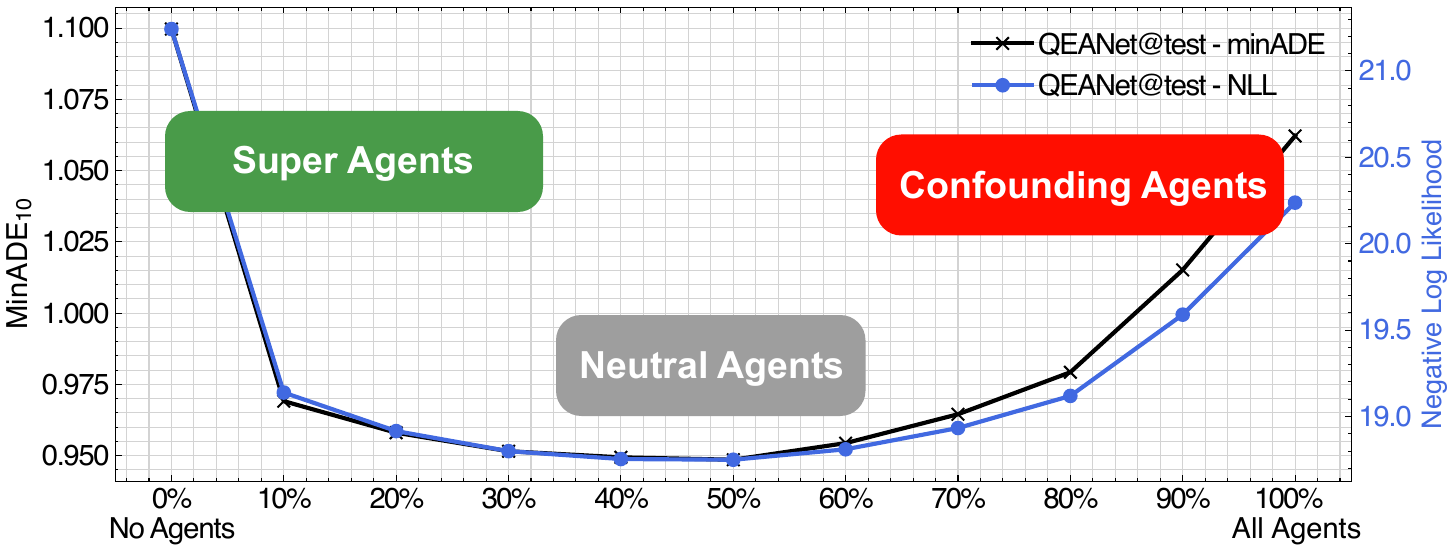}
  
        \caption{\textit{Insertion Test} results.}

        \label{fig:page_one_fig_bottom}
    \end{subfigure}
    
    \caption{(a) Removing a Confounding Agent can markedly improve prediction accuracy. (b) The Insertion Test reveals a U-shaped curve: using only the most helpful agents yields the best results.}
\end{figure}
Our contribution in this work is twofold. As a primary contribution, we adopt Shapley-based attribution methods \cite{shapley7ValueNPerson2020, makansiYOUMOSTLYWALK2022} to quantify the contribution of each surrounding agent to prediction performance. This serves as a post-hoc evaluation method leveraging ground-truth data. Building on \cite{roelofsCausalAgentsRobustnessBenchmark2022}, which demonstrated that removing certain groups of agents can sometimes even improve predictions, we extend the analysis to the individual-agent level. We uncover a striking imbalance among surrounding agents. A small subset of agents, which we refer to as \textit{Super Agents}, enhances prediction accuracy. In contrast, a larger subset, termed \textit{Confounding Agents}, introduces spurious or misleading information that reduces model performance, as depicted in Fig.~\ref{fig:page_one_fig_bottom}. 
The positive and negative contributions nearly cancel out, leaving only marginal performance gains compared to model performance when ignoring surrounding agents altogether. Notably, this behavior is observable across different metrics. This analysis provides a principled explanation for the lack of robustness observed in practice \cite{caoAdvDORealisticAdversarial2022, zhangAdversarialRobustnessTrajectory2022, roelofsCausalAgentsRobustnessBenchmark2022} and highlights the sensitivity of current models to spurious features in contextual information.

Second, we explore the application of the Conditional Information Bottleneck (CIB) approach \cite{alemiDeepVariationalInformation2019, leeConditionalGraphInformation2023} to filter out non-robust agent information. We find that constraining the information flow consistently improves robustness across various models and datasets. 

Our contributions can be summarized as follows:
\begin{itemize}
\item A systematic analysis of state-of-the-art trajectory prediction models, revealing their lack of robustness when incorporating surrounding agent information. 
\item Evidence that prediction accuracy is improved by only a small subset of Super Agents, whereas many agents act as confounders that degrade performance; including a comparison with causality-based relevancy labels.
\item A Conditional Information Bottleneck (CIB) module that improves robustness by filtering spurious agent information without additional supervision, augmented by a comprehensive analysis of its benefits and limitations.
\end{itemize}

\section{Related Work}
\subsection{Trajectory Prediction}
Accurate trajectory prediction is essential for autonomous driving and robotics, enabling systems to anticipate the future movements of surrounding agents. Recent methods fall into two main categories: marginal prediction and joint prediction.
Marginal prediction forecasts the trajectory of a target agent conditioned on contextual cues like the lane graph and nearby agents. State-of-the-art models such as LAformer \cite{liuLAformerTrajectoryPrediction2023}, QEANet \cite{chenQEANetImplicitSocial2024}, and other transformer-based methods \cite{shiMotionTransformerGlobal2023, linEDAEvolvingDistinct2023} set benchmarks on datasets like nuScenes \cite{caesarNuScenesMultimodalDataset2020} and Waymo Open Motion Dataset (WOMD) \cite{ettingerLargeScaleInteractive2021}.

In contrast, joint prediction forecasts trajectories for multiple agents simultaneously to capture social interactions. While mechanisms like cross-attention \cite{zhouQueryCentricTrajectoryPrediction2023, zhouQCNeXtNextGenerationFramework2023, wangSimpleMultiagentJoint2024} are common, recent advances such as BeTop \cite{liuReasoningMultiAgentBehavioral} introduce Behavioral Topology to explicitly represent consensual behavioral patterns. We benchmark baselines from both paradigms.

\subsection{Quantifying Feature Influence}
For safety critical tasks, uncovering the decision-making mechanisms of black-box models is crucial to ensure reliable behavior in complex and unseen scenarios. But quantifying the influence of surrounding agents remains a challenging task due to the absence of ground truth labels. Prior work addresses \cite{roelofsCausalAgentsRobustnessBenchmark2022} the lack of ground-truth labels for agent influence by manually annotating agents as causal or non-causal. Their analysis based on removing these agents found that models are highly sensitive to non-causal ones due to spurious correlations. While a proposed training-time augmentation involving dropping these agents improved robustness, it requires prohibitively large labeling efforts for big datasets and is subjective.
We generalize these findings across different models \cite{chenQEANetImplicitSocial2024, liuLAformerTrajectoryPrediction2023, shiMTRMultiAgentMotion2024, linEDAEvolvingDistinct2023} and datasets \cite{caesarNuScenesMultimodalDataset2020, ettingerLargeScaleInteractive2021}, and propose a plug-and-play solution that does not require additional labels or changes in the training setup.

In \textit{You Mostly Walk Alone} \cite{makansiYOUMOSTLYWALK2022}, Makansi et al. apply a Shapley value-based analysis to reveal that a target agent’s past trajectory dominates predictions, while interactions contribute little. By inserting random dummy agents, which received attribution scores nearly identical to real neighbors, they demonstrated that models often ignore relevant social context. Building on this, we use Shapley values for a systematic investigation across recent models. Unlike \cite{makansiYOUMOSTLYWALK2022}, we find that performance is limited not by a total absence of neighbor attention, but by the inconsistent influence of impactful agents, whose positive and negative contributions often cancel each other out.  

The limitations of standard interpretability tools, such as Transformer attention weights or gradient-based saliency maps, further justify the need for more robust attribution methods. While these mechanisms are often used to identify influential features, research indicates that they are frequently insufficient for explanation. Jain et al. \cite{jainAttentionNotExplanation2019} demonstrate that attention weights often do not correlate with feature importance metrics. Furthermore, Adebayo et al. \cite{adebayoSanityChecksSaliency2018} highlight that many saliency methods fail basic sanity checks, acting as simple edge detectors that remain independent of both model parameters and the data-generating process. 

These observations motivate our Shapley-based analysis and the exploration of the \textit{Conditional} Information Bottleneck (CIB) \cite{leeConditionalGraphInformation2023} as a principled approach to filtering spurious agent information and improving robustness.

\subsection{Focusing on relevant features}

In the field of motion prediction for autonomous driving, recent works seek to improve robustness and generalization by reducing the influence of non-causal agents.
Causal trajectory prediction (CRiTIC) \cite{ahmadiCurbYourAttention2025} proposes a Causal Discovery Network (CDN) trained to extract causal links between agents from their past trajectories. The CDN generates a causal graph that guides prediction, and a sparsity regularization loss along with a self-supervised auxiliary task suppresses information from agents classified as non-causal. Similarly, Pourkeshavarz et al. \cite{pourkeshavarzCaDeTCausalDisentanglement2024} propose a causal disentanglement framework (CaDeT). Their method separates invariant (causal) from variant (spurious) features by training on an intervention set created from the measured uncertainty statistics in the latent space. Spurious features are replaced with samples from the intervention set, and the model is trained to remain invariant to these substitutions. Both draw on information-bottleneck principles \cite{tishbyInformationBottleneckMethod2000} and report robustness gains.

Building on this line of work, and complementing our Shapley-value-based agent attribution analysis, we evaluate the use of the CIB across multiple state-of-the-art architectures. We perform an ablation study comparing model behavior with and without the CIB and examine how the CIB affects the decision-making process.

\section{Preliminaries and Methodology}
\label{section:preliminaries}
To systematically analyze how trajectory prediction models use information from surrounding agents, we first use Shapley-based attribution methods to quantify each agent's individual contribution to prediction accuracy. We propose to classify agents with a performance-improving attribution as Super Agents. The prediction performance given only the Super Agents is a key aspect of our work. Based on the individual agent attributions, we propose to perform an \textit{insertion test} (see later) to evaluate how model performance varies when considering different subsets of agents. We then propose to implement the Conditional Information Bottleneck as a potential method to mitigate the negative impact of non-beneficial agent information revealed by our analysis. The analysis is then reused to verify the CIB's effectiveness.

\subsection{Attributing performance to features}\label{sec:attribution_methods}
Feature attribution methods in Explainable AI \cite{makansiYOUMOSTLYWALK2022} quantify how each input feature contributes to a model's prediction. This analysis is performed post-hoc, requiring ground-truth data to evaluate feature importance with respect to a specific performance metric.
Shapley values \cite{shapley7ValueNPerson2020} iterate over all possible combinations of features in order to compute the marginal contribution each individual feature has to the prediction. 
The Shapley value $\phi_i$ for a feature $i$ is given by:

\begin{equation}
\phi_i(v) = \sum_{S \subseteq N \setminus \{i\}} \frac{|S|!(|N|-|S|-1)!}{|N|!} (v(S \cup \{i\}) - v(S))
\end{equation}
where:
\begin{itemize}
    \item $N$ is the set of all features,
    \item $S$ is a subset of features not including feature $i$,
    \item $v(S)$ is the value function (e.g., model prediction) for the coalition of features in set $S$,
    \item $v(S \cup \{i\}) - v(S)$ is the marginal contribution of feature $i$ to coalition $S$.

\end{itemize}
In this work, the value function \( v \) depends on both the underlying model and a target metric \( m(\cdot) \), such that \( v(S) = m(f(S)) \), where \( f(S) \) is the model's output when only features in \( S \) are used. We largely focus on the negative-log-likelihood (NLL), since it considers all aspects of the prediction.

Computing exact Shapley values is expensive, scaling exponentially with the number of features ($2^n$ evaluations), which is impractical for large input sets. To address this, ApproShapley \cite{castroPolynomialCalculationShapley2009} offers an efficient approximation by randomly sampling a subset of feature permutations.

To evaluate the faithfulness of attribution methods, Petsiuk et al. \cite{petsiukRISERandomizedInput2018} introduced \textit{insertion and deletion tests} for image classification. This approach assesses the attributions' quality by sequentially adding (inserting) or removing (deleting) features according to their importance and observing the effect on the model's output. A steep performance increase during insertion or a sharp decline during deletion indicates a more faithful attribution. Hama et al. \cite{hamaDeletionInsertionTests2023} later extended this evaluation to regression models.
Since this work focuses on the handling of the surrounding agent information, we adapt the aforementioned methodology to ablate individual surrounding agents. By sequentially inserting agents based on their attribution scores, we investigate the quantity and influence of helpful, irrelevant, and deteriorating agents.

\subsection{Intra- and Inter-Model Agreement}

Let $\phi_{i,j,\text{NLL}}$ be the Shapley value of Agent $i$ for the $j$-th model regarding the NLL metric. The rate $r_{i,\text{NLL}}$ represents the agreement of $N$ models on an agent $i$ being helpful ($\phi_{i,\text{NLL}} < 0$) to the prediction performance. Defined as the mean of an indicator function $\mathds{1}(\phi_{i,j,\text{NLL}} < 0)$ over $N$ total models for each agent $i$:
\begin{equation}
r_{i,\text{NLL}} = \frac{1}{N} \sum_{j=1}^{N}  \mathds{1} (\phi_{i,j,\text{NLL}} < 0).    
\end{equation}
We propose to apply this analysis in the subsequent sections to assess attributional agreement across models. A value of $r_{i,\text{NLL}} = 1$ indicates consistent attribution with beneficial influence to agent $i$, whereas $r_{i,\text{NLL}} = 0$ shows consistent identification of the agent as confounding. Intermediate values ($0 < r_{i,\text{NLL}} < 1$) capture ambiguous cases, reflecting divergences in the decision-making mechanisms of different models. This agreement-based perspective enables us to distinguish agents that are universally informative from those whose attributed importance is unstable and model-dependent.
We distinguish between \textit{intra-}model agreement, where the same model is evaluated using different inference seeds, and \textit{inter-}model agreement, where independently trained models of the same architecture are evaluated.

\subsection{Attribution-based Performance}
To analyze the contribution of surrounding agents, we evaluate model performance under three distinct conditions: (i) using All agents (default configuration), (ii) using only Super Agents, and (iii) No agents. 
The set of Super Agents comprises all agents in a scene whose attribution indicates a beneficial influence on prediction performance i.e., a negative attribution with respect to the NLL metric. Formally,
\begin{align}
\mathcal{A}_{\text{Super}} = \{ i \in \mathcal{A}_{\text{All}} \mid \phi_{i,\text{NLL}} < 0 \},
\end{align}
The comparison between the three settings disentangle the effects of Super versus Confounding Agents. We define two performance gaps with respect to a metric $m(\cdot)$:
\begin{align}
\Delta^{m}_{\text{Super-All}} &= m(\text{Super}) - m(\text{All}) \\
\Delta^{m}_{\text{No-All}} &= m(\text{No}) - m(\text{All}) 
\end{align}
The gap $\Delta^{m}_{\text{Super-All}}$ captures the negative impact of including uninformative or confounding agents. Ideally, this gap should approach zero if all agents contribute positively or neutrally. In contrast, $\Delta^{m}_{\text{No-All}}$ quantifies the benefit of leveraging surrounding agents relative to using none.


\subsection{Information Bottleneck}

The Information Bottleneck (IB) principle, introduced by Tishby et al. \cite{tishbyInformationBottleneckMethod2000}, provides a theoretical framework for extracting the most relevant information from an input variable \( X \) with respect to a target variable \( Y \). The goal is to learn a compressed representation \( T \) of \( X \) that preserves information required for predicting \( Y \), while discarding irrelevant details. Formally, this trade-off is achieved by minimizing
\begin{align}
    \mathcal{L}_{\text{IB}} = -I(Y; T) + \beta I(X; T),
\end{align}

where \( I(\cdot;\cdot) \) denotes mutual information and \( \beta \) is a Lagrange multiplier that controls the balance between compression and predictive power. The IB framework has been extended to deep learning through the Variational Information Bottleneck (VIB) \cite{alemiDeepVariationalInformation2019}. In graph learning, for example, IB has been adapted to identify minimal yet informative substructures, allowing the model to focus on task-relevant subgraphs while suppressing redundancy in the input graph topology \cite{yuImprovingSubgraphRecognition2022,leeConditionalGraphInformation2023}.

\subsubsection*{Conditional Information Bottleneck}
While the standard Information Bottleneck (IB) framework considers compression of an input variable with respect to a target, Lee et al.~\cite{leeConditionalGraphInformation2023} extend this principle to the \textit{Conditional Information Bottleneck} (CIB) framework, which addresses settings where the relevance of information depends on a given context. Given input variables \( X_1 \), \( X_2 \), and a target variable \( Y \), the CIB framework seeks a compressed representation \( T_1 \) of \( X_1 \) that preserves information about \( Y \) conditioned on \( X_2 \). This is formalized by minimizing the Conditional Information Bottleneck objective:

\begin{align}
\mathcal{L}_{\text{CIB}} = -I(Y; T_1 \mid X_2) + \beta I(X_1; T_1 \mid X_2),
\end{align}
where \( I(\cdot;\cdot \mid \cdot) \) denotes conditional mutual information, and \(\beta\) controls the trade-off between compressing \( X_1 \) and preserving task-relevant information for predicting \( Y \) given \( X_2 \). 

In essence, the CIB objective encourages \( T_1 \) to retain only those aspects of \( X_1 \) that are informative for predicting \( Y \) given \( X_2 \), while discarding irrelevant or redundant information. This conditional perspective is particularly useful in multi-view or paired-input settings, such as learning from paired graphs, where one graph provides context for interpreting the other \cite{federiciLearningRobustRepresentations2020, leeConditionalGraphInformation2023}. In autonomous driving, this is highly applicable: to predict a target agent's trajectory \( Y \) from its own kinematic history \( X_2 \), we condition on the information of relevant surrounding agents \(X_1\). The CIB framework would learn a representation of the surrounding agents \(T_1\) that emphasizes relevant movements with respect to the target agent's movement. 

All models in our study follow an Encoder-Interactor-Decoder structure \cite{chenQEANetImplicitSocial2024, liuLAformerTrajectoryPrediction2023, shiMTRMultiAgentMotion2024, linEDAEvolvingDistinct2023}. To investigate the effect of conditional information compression, we augment each model with a CIB module. The CIB is inserted after the encoding of the surrounding agent information and is conditioned on the encoding of the target agent as shown in Fig. \ref{fig:architecture_cib}.

\begin{figure}[ht]
\centering
\resizebox{0.45\textwidth}{!}{
\begin{tikzpicture}[
    block/.style={
        rectangle, 
        draw=black, 
        fill=blue!20, 
        text=black, 
        font=\sffamily\bfseries, 
        align=center, 
        minimum width=3.2cm, 
        minimum height=1.8cm, 
        rounded corners=10pt,
        line width=0.8pt
    },
    cib_block/.style={
        block,
        fill=gray!10
    },
    input_label/.style={
        font=\sffamily\large\itshape,
        align=center,
        inner sep=2pt
    },
    arrow/.style={
        -{Stealth[scale=1.0]},
        draw=black,
        line width=1pt
    },
    dashed_arrow/.style={
        arrow,
        dashed
    }
]

    \node[block] (surr) {SurrAgent\\Encoder};
    \node[block, below=0.8cm of surr] (target) {Target Agent\\Encoder};
    \node[block, below=0.8cm of target] (lane) {Lane Graph\\Encoder};
    
    \node[cib_block, right=1.2cm of surr] (cib) {CIB};
    \node[block, right=1.2cm of target] (interactor) {Interactor};
    
    \node[block, right=1.0cm of interactor] (decoder) {Decoder};

    \node[input_label, left=0.4cm of surr] (surr_in) {$\{x, y, v_x, v_y, w, l, \theta\}_s$};
    \node[input_label, left=0.4cm of target] (target_in) {$\{x, y, v_x, v_y, w, l,  \theta\}_t$};
    \node[input_label, left=0.4cm of lane] (lane_in) {Polyline nodes, \\ Central lines};

    \draw[arrow] (surr_in) -- (surr);
    \draw[arrow] (target_in) -- (target);
    \draw[arrow] (lane_in) -- (lane);

    \draw[arrow] (surr) -- (cib);
    \draw[arrow] (target) -- (interactor);
    \draw[arrow] (interactor) -- (decoder);
    \draw[arrow] (cib) -- (interactor);
    
    \draw[arrow] (lane.east) -- (interactor.south);
    
    \draw[dashed_arrow] (target) -- (cib);

\end{tikzpicture}
} 
\caption{Proposed architecture. The CIB module compresses surrounding-agent features while conditioning on the target agent's state; the result is fused with lane geometry and target embeddings in the Interactor before decoding.}
\label{fig:architecture_cib}
\end{figure}

\section{Experiments}
Our experimental investigations focus on two widely recognized and comprehensive autonomous driving datasets: Waymo Open Motion Dataset (WOMD) \cite{ettingerLargeScaleInteractive2021} and nuScenes \cite{caesarNuScenesMultimodalDataset2020}. Both datasets offer rich multi-modal sensor data, detailed annotations of traffic participants, and high-definition maps, providing a robust foundation for evaluating motion prediction models. 
For the marginal prediction task, we investigate two state-of-the-art models for each dataset. For nuScenes, we evaluate QEANet \cite{chenQEANetImplicitSocial2024} and LAformer \cite{liuLAformerTrajectoryPrediction2023}. For WOMD, we evaluate MTR \cite{shiMotionTransformerGlobal2023} and EDA \cite{linEDAEvolvingDistinct2023}, using 20\% of the dataset to reduce computational cost.

To estimate variability, each model variant is trained with five random seeds, and we report the mean and standard deviation across seeds.

In addition to that, we assess BeTop \cite{liuReasoningMultiAgentBehavioral} within the context of the Waymo Interaction Challenge for joint motion forecasting. To evaluate the impact of our proposed method, we train both the base BeTop model and a CIB-enhanced variant across three training seeds on the complete Waymo dataset.

For CIB-enhanced variants, the bottleneck weight $\beta$ was optimized via grid search in the range $[10^{-2}, 10^{2}]$.
For evaluation, we adopt standard trajectory prediction metrics \cite{chenQEANetImplicitSocial2024, liuLAformerTrajectoryPrediction2023, shiMotionTransformerGlobal2023, linEDAEvolvingDistinct2023}. Average Displacement Error (ADE) measures the mean Euclidean distance between predicted and ground-truth trajectories, while Final Displacement Error (FDE) captures the error at the final timestep. Since models produce multiple trajectories, we report minADE/minFDE@K, which reflects the best among $K$ predictions, and Miss Rate, the percentage of cases where none of the $K$ trajectories lie within a threshold distance of the ground truth at the final timestep. To assess the quality of probabilistic predictions, the Negative Log-Likelihood (NLL) of the ground truth trajectory is measured.
\footnote{
Direct comparison between nuScenes and WOMD metrics is not possible due to differing evaluation protocols: nuScenes reports errors at a 6-second horizon, whereas WOMD averages across multiple horizons (e.g., 3, 5, 8 seconds).
}

\subsection{Robustness}
We evaluate model robustness using two types of perturbations: removal-based and noise-based.

The removal-based perturbations are introduced in \cite{roelofsCausalAgentsRobustnessBenchmark2022}. Using agents labeled as causal or non-causal \cite{roelofsCausalAgentsRobustnessBenchmark2022}, we analyze model behavior when removing either causally relevant or Non-Causal Agents. We apply this perturbation to the WOMD-based MTR~\cite{shiMotionTransformerGlobal2023} and EDA~\cite{linEDAEvolvingDistinct2023}.
To complement the targeted removal perturbation, we introduce a noise-based perturbation that emulates sensor noise and perception errors common in real-world driving. By applying Gaussian noise to the input trajectories, we can measure the model's stability and robustness to a different type of perturbation. We apply this perturbation to the nuScenes-based models: QEANet~\cite{chenQEANetImplicitSocial2024} and LAformer~\cite{liuLAformerTrajectoryPrediction2023}.
To quantify the sensitivity to these perturbations, we use the evaluation metric from \cite{roelofsCausalAgentsRobustnessBenchmark2022}. Specifically, their investigations are based on the absolute deviation from the original performance: 

\begin{align}
\begin{split}
    Abs(\Delta) = \frac{1}{n} \sum^n_{i=1} \Biggl| & \minADE(f(x_{i,\text{perturbed}}), y_i) \\
    & - \minADE(f(x_{i,\text{original}}), y_i) \Biggr|
\end{split}
\end{align}

where $n$ is the total number of samples, $f$ is the prediction model, $x_{i,\text{original}}$ is the $i$-th original input scene, $x_{i,\text{perturbed}}$ is its corresponding perturbed version, $y_i$ is the ground truth trajectory.

\begin{figure}[htbp]
    \centering

    \includegraphics[width=0.92\linewidth, trim={0cm 0cm 0cm 0cm}, clip]{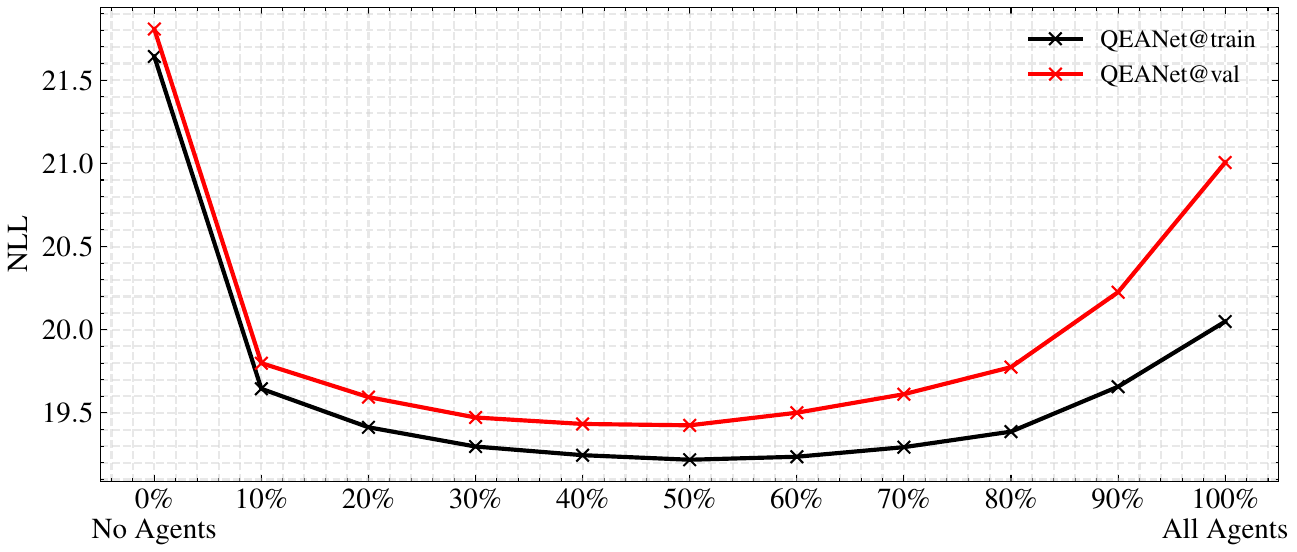}
        \caption{Insertion test on the nuScenes train and validation sets. On validation, Confounding Agents have a much stronger effect, almost canceling out the Super Agents.}
        \label{fig:queanet_insertion_test}

\end{figure}

\begin{figure}[t] 
    \centering
    \begin{subfigure}[b]{0.48\columnwidth} 
        \centering
        \includegraphics[width=1\linewidth]{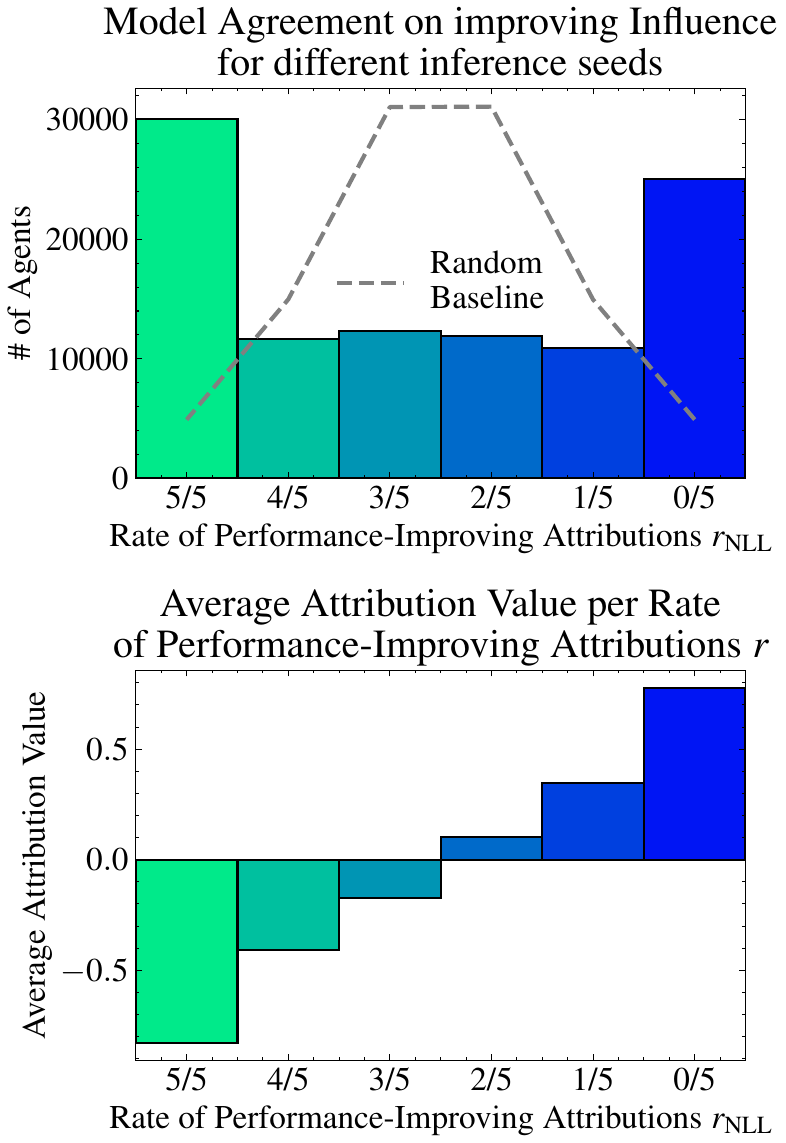}
        \caption{Intra-Model}
        \label{fig:intra_model_agreement}
    \end{subfigure}
    \hfill 
    \begin{subfigure}[b]{0.48\columnwidth}
        \centering
        \includegraphics[width=1\linewidth]{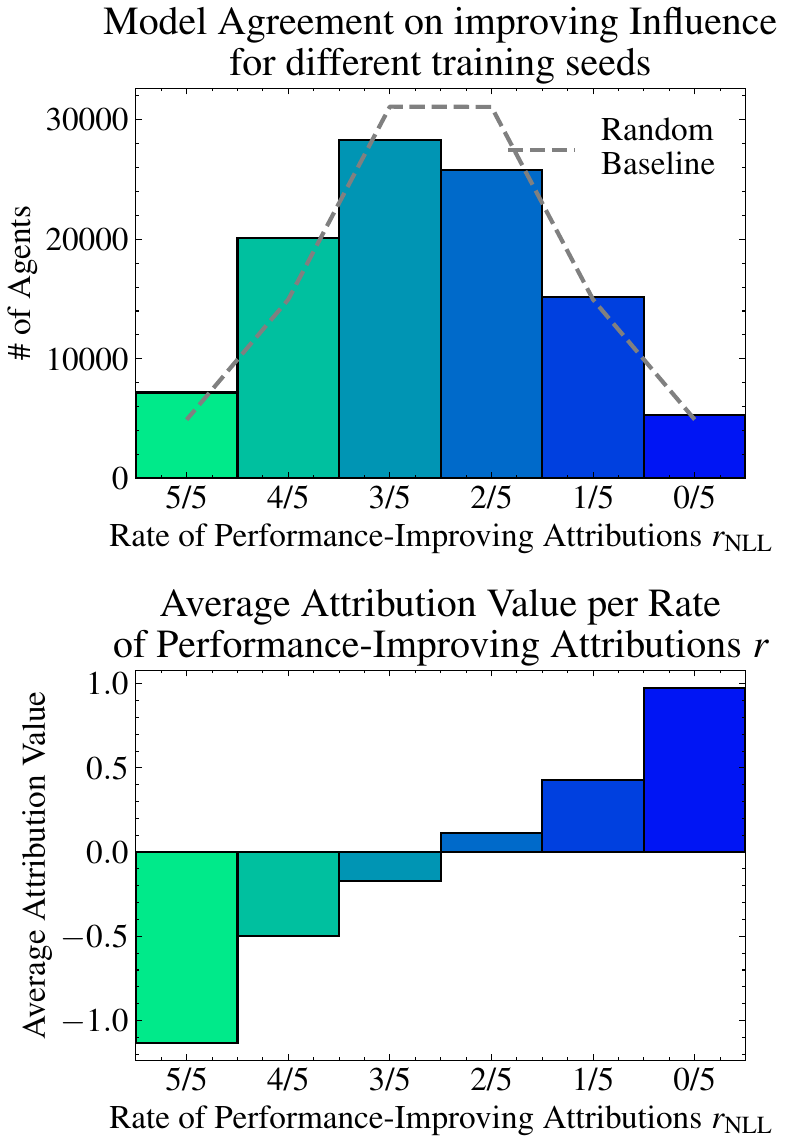}
        \caption{Inter-Model}
        \label{fig:inter_model_agreement}
    \end{subfigure}
    
    \caption{Consistency of agent influence for QEANet on nuScenes. (a) Intra-model: a single model is highly consistent across five inference seeds. (b) Inter-model: models trained with different seeds vary significantly, following the random baseline, indicating unstable learned decision-making.}
    \label{fig:combined_agreement}
\end{figure}

\begin{figure*}[t]
    \centering
    \begin{subfigure}[b]{0.24\textwidth}
        \includegraphics[width=1\linewidth, trim={0cm 0cm 0cm 0cm}, clip]{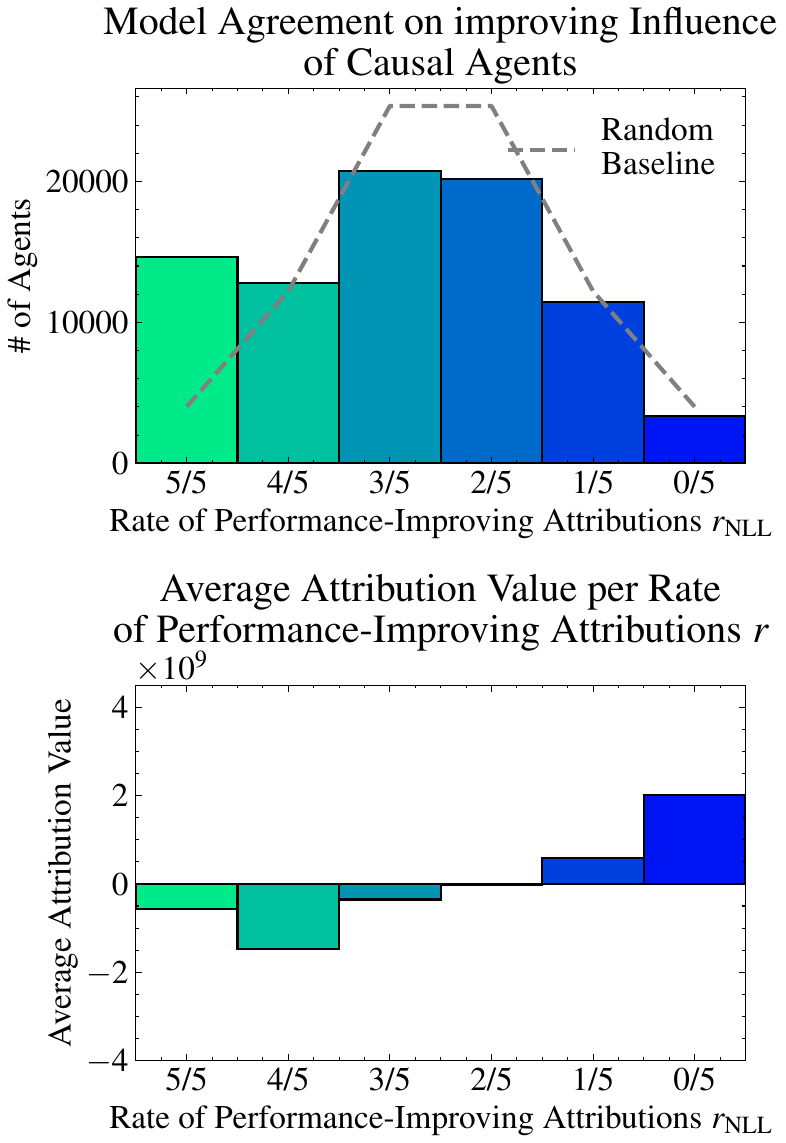}
        \caption{Causal Agents \\ on MTR}
        \label{fig:inter_model_agreement_mtr_base_causal}
    \end{subfigure}
    \hfill
    \begin{subfigure}[b]{0.24\textwidth} 
        \includegraphics[width=1\linewidth, trim={0cm 0cm 0cm 0cm}, clip]{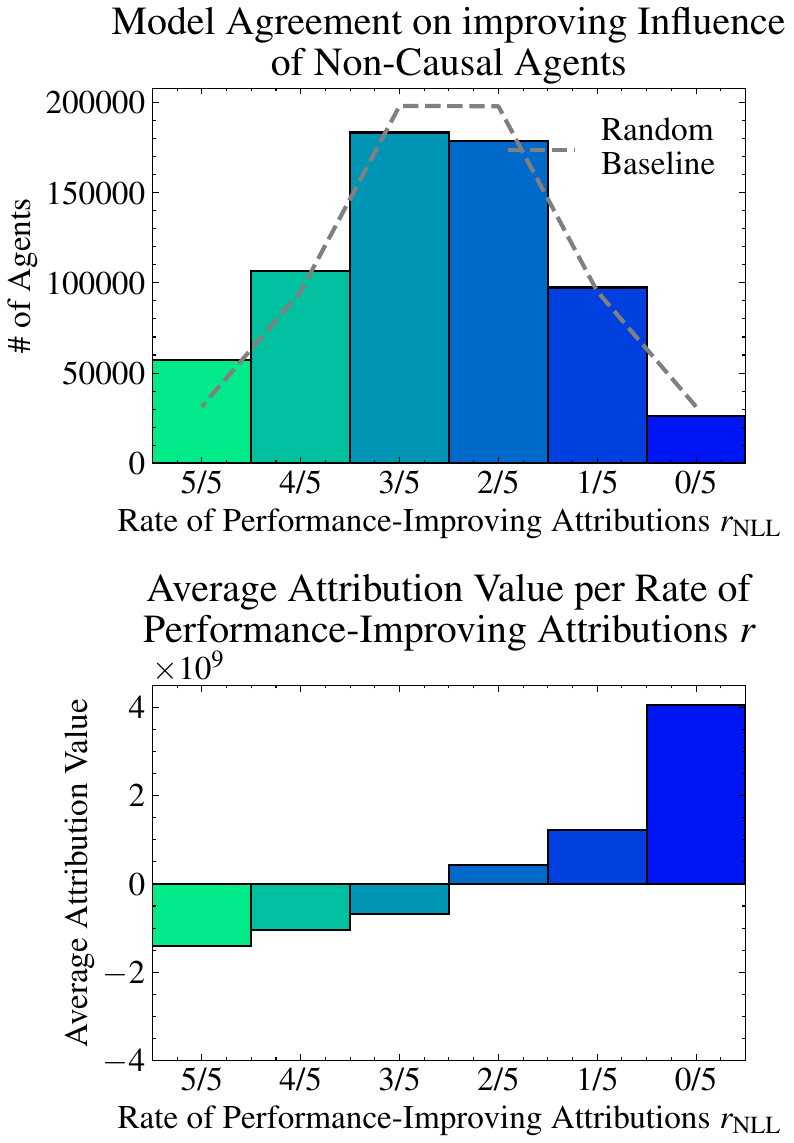}
        \caption{Non-causal Agents \\ on MTR}
        \label{fig:inter_model_agreement_mtr_base_noncausal}
    \end{subfigure}
    \hfill
    \begin{subfigure}[b]{0.24\textwidth} 
        \includegraphics[width=1\linewidth, trim={0cm 0cm 0cm 0cm}, clip]{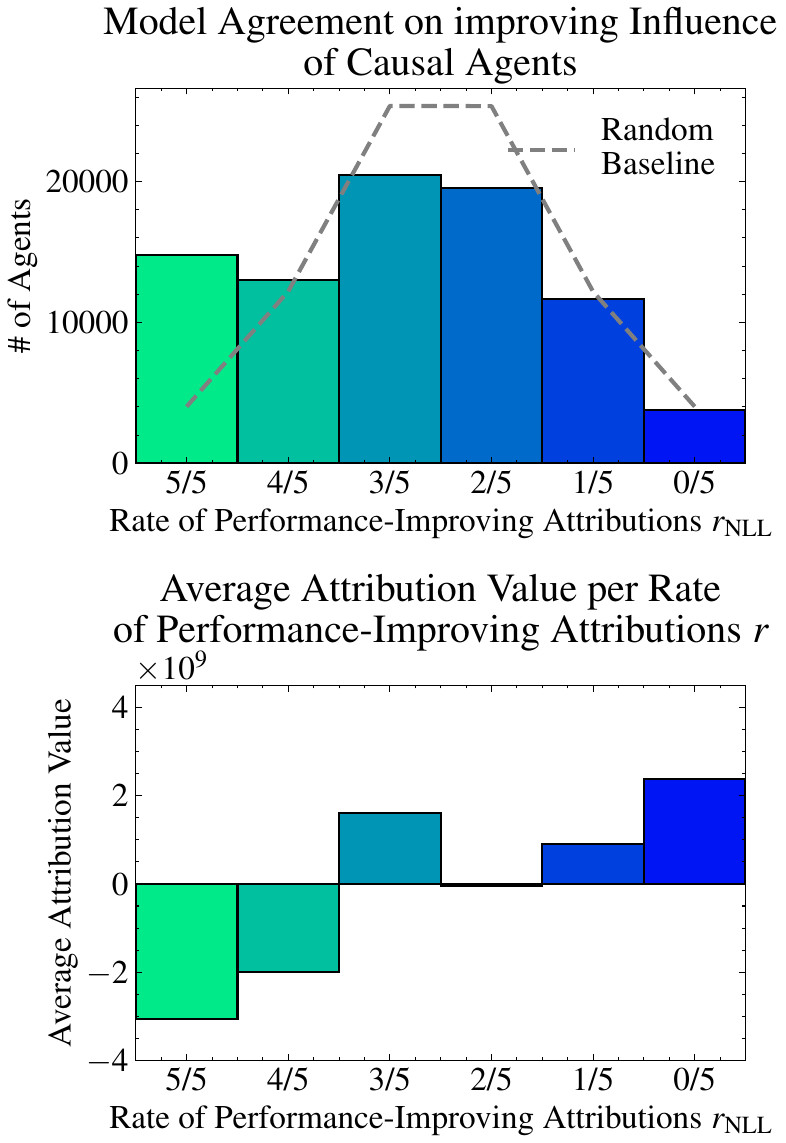}
        \caption{Causal Agents \\ on MTR+CIB}
        \label{fig:inter_model_agreement_mtr_ib_causal}
    \end{subfigure}
    \hfill
    \begin{subfigure}[b]{0.24\textwidth} 
        \includegraphics[width=1\linewidth, trim={0cm 0cm 0cm 0cm}, clip]{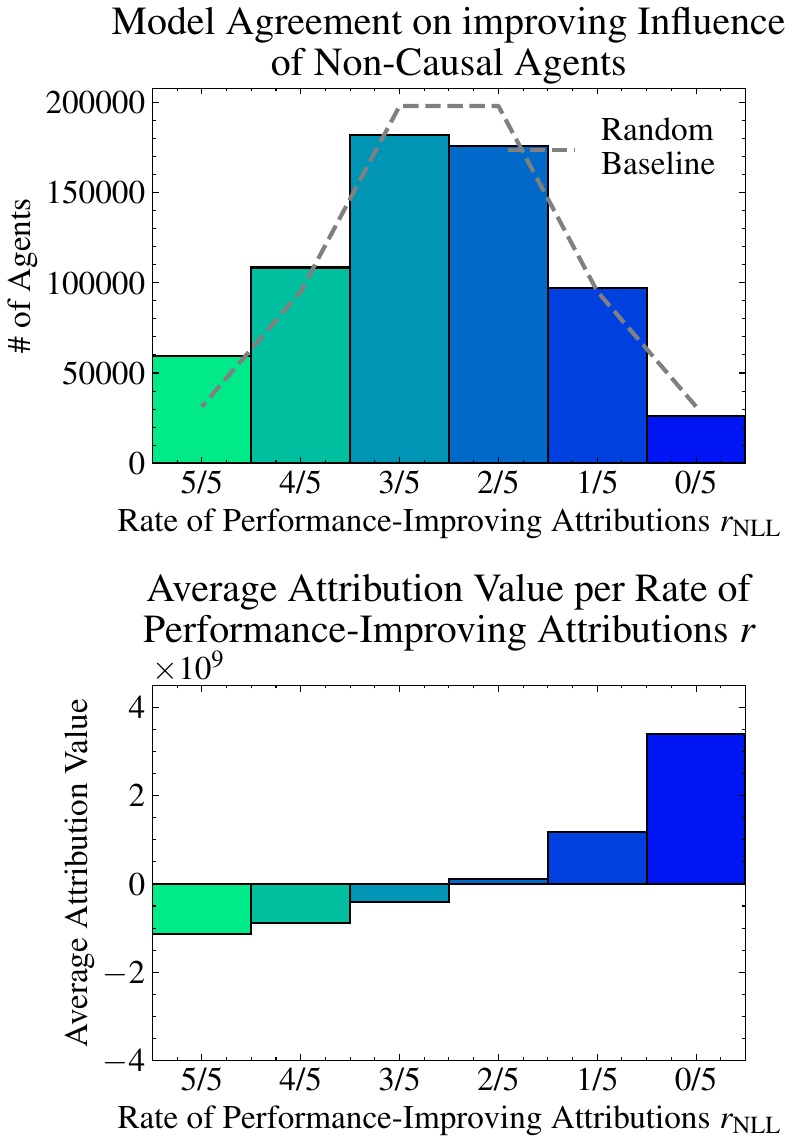}
        \caption{Non-causal Agents \\     on MTR+CIB}        
        \label{fig:inter_model_agreement_mtr_ib_noncausal}
    \end{subfigure}
    \caption{
    Consistency of improving agent influence for the MTR model and the MTR+CIB model compared to human labels. (a) Causal Agents are only slightly more likely to be performance-improving. (b) Non-Causal Agents follow the random baseline, showing the model cannot identify them. For MTR+CIB: (c) the number of improving Causal Agents does not rise, but the mean attribution at $r_{\text{NLL}}=\frac{5}{5}$ drops significantly; (d) Non-Causal usage still matches the random baseline.}
    \label{fig:combined_causality_agreement}
\end{figure*}

\begin{table*}[t]
\centering
\begin{tabular}{ | l | c | c | c | c | c |}
 \hline
 Model & $\mathrm{minADE}_{10} \downarrow $ & $\mathrm{minADE}_{5} \downarrow $ & $\mathrm{MissRate}_{10} \downarrow $ & $\mathrm{MissRate}_{5} \downarrow $ & $ \mathrm{NLL} \downarrow$\\
 \hline
 \rowcolor{gray!25} QEANet & $1.053 \pm 0.010$ & $1.203 \pm 0.012$ & $0.479 \pm 0.009$ & $0.511 \pm 0.009$ & $20.125 \pm 0.210$ \\
 - Super Agents & $0.961 \pm 0.012$ & $1.106 \pm 0.012$ & $0.443 \pm 0.009$ & $0.475 \pm 0.009$ & $18.336 \pm 0.276$ \\
 - No Agents  & $1.116 \pm 0.014$ & $1.317 \pm 0.012$ & $0.542 \pm 0.010$ & $0.574 \pm 0.008$ & $21.025 \pm 0.253$ \\
 \rowcolor{gray!25} QEANet + CIB & $\mathbf{1.042 \pm 0.016}$ & $\mathbf{1.191 \pm 0.017}$ & $\mathbf{0.471 \pm 0.008}$ & $\mathbf{0.504 \pm 0.008}$ & $\mathbf{19.820 \pm 0.341}$ \\
  - Super Agents & $0.977 \pm 0.017$ & $1.122 \pm 0.013$ & $0.448 \pm 0.010$ & $0.479 \pm 0.006$ & $18.629 \pm 0.338$ \\
  - No Agents & $1.109 \pm 0.016$ & $1.316 \pm 0.018$ & $0.538 \pm 0.005$ & $0.571 \pm 0.002$ & $20.868 \pm 0.328$ \\
 \hline
 
 \rowcolor{gray!25} LAformer & $\mathbf{0.948 \pm 0.012}$ & $\mathbf{1.604 \pm 0.018}$ & $\mathbf{0.348 \pm 0.003}$ & $\mathbf{0.504 \pm 0.007}$ & $36.914 \pm 1.462$ \\
 - Super Agents & $0.882 \pm 0.012$ & $1.560 \pm 0.018$ & $0.326 \pm 0.007$ & $0.483 \pm 0.002$ & $32.150 \pm 0.982$ \\
 - No Agents  & $1.024 \pm 0.013$ & $1.913 \pm 0.042$ & $0.406 \pm 0.004$ & $0.556 \pm 0.004$ & $38.461 \pm 2.130$ \\
\rowcolor{gray!25} LAformer + CIB & $0.985 \pm 0.006$ & $1.627 \pm 0.020$ & $0.380 \pm 0.007$ & $\mathbf{0.504 \pm 0.004}$ & $\mathbf{35.741 \pm 0.830}$ \\
  - Super Agents  & $0.924 \pm 0.007$ & $1.592 \pm 0.037$ & $0.367 \pm 0.007$ & $0.498 \pm 0.005$ & $31.412 \pm 0.637$ \\
  - No Agents & $1.032 \pm 0.012$ & $1.810 \pm 0.061$ & $0.425 \pm 0.009$ & $0.548 \pm 0.008$ & $34.029 \pm 1.220$ \\
\hline

\end{tabular}
\caption{QEANet and LAformer trained and evaluated on nuScenes. No Agents is the baseline without social context; Super Agents uses only improving-influence agents. The three rows correspond to the right endpoint, minimum, and left endpoint of the insertion curve (Fig.~\ref{fig:page_one_fig_bottom}), respectively. Note that Super Agent identification is a post-hoc, model-dependent analysis relative to a specific metric (NLL here) and requires ground truth. Results are averaged over five runs (mean $\pm$ std).}
\label{table:results_performance_nuscenes}
\end{table*}

\section{Results}
This section first analyzes how models utilize surrounding agents via the Insertion Test, followed by an investigation into Model Agreement and alignment with Causal Agents. Finally, we assess Benchmark Performance and Robustness across multiple architectures to quantify the CIB's impact.
\subsection{Insertion Test}

Based on the insertion test introduced in \ref{sec:attribution_methods}, we analyze the handling of surrounding agent information in QEANet. Fig.~\ref{fig:queanet_insertion_test} shows that a small subset of agents is beneficial (Super Agents), while many others degrade performance (Confounding Agents), 
an effect that is more pronounced on the validation set compared to the train set, which hints at overfitting on irrelevant information.

These observations offer a key insight: while the model successfully uses Super Agents, it struggles to ignore Confounding Agents. This leads us to investigate whether the model's Super Agents align with the agents that are causally relevant in a scene. If this alignment exists, the model's internal attribution values could potentially be leveraged as a powerful supervisory signal to guide and improve its relevancy predictions during training.

\subsection{Intra- and Inter-Model Agreement}
To account for the probabilistic nature of the predictor, we calculate the attribution $\phi_{i,\text{NLL}}$ for each agent $i$ on the same model on five different inference seeds (Fig.~\ref{fig:intra_model_agreement}) and for five models trained on different training seeds (Fig.~\ref{fig:inter_model_agreement}). 
Based on the frequency of $\phi_{i,\text{NLL}} < 0$ for each agent $i$ in different settings, we can better understand the decision-making process.

Fig.~\ref{fig:intra_model_agreement} shows that, within the same model and across inference seeds, most agents consistently either improve or degrade performance.
Only a small number fall into an ambiguous middle range, indicating that individual agent influence is relatively consistent under different inference conditions. In contrast, Fig.~\ref{fig:inter_model_agreement} illustrates that models trained with different seeds disagree substantially on which agents are helpful and follow closely the Random Baseline indicator. Despite sharing the same architecture and training data, the models learn divergent decision-making patterns. This discrepancy suggests that the model does not robustly capture invariant relationships between agents, contradicting the expectation that models trained multiple times on the same dataset would identify the same decision-making mechanisms.

We further analyze how Shapley attributions correspond to human-labeled Causal Agents from the CausalAgents benchmark \cite{roelofsCausalAgentsRobustnessBenchmark2022}. Figure~\ref{fig:inter_model_agreement_mtr_base_causal} shows that Causal Agents exhibit a tendency toward negative Shapley values, indicating a weak performance-improving effect. Many Causal Agents are attributed with positive Shapley values, which degrade performance, indicating unfaithful decision-making.

Complementing this finding, the influence distribution for Non-Causal agents (Fig.~\ref{fig:inter_model_agreement_mtr_base_noncausal}) closely follows the random baseline. This confirms that the model processes surrounding agent information without prioritizing causally relevant actors. 

Fig.~\ref{fig:inter_model_agreement_mtr_ib_causal} and Fig.~\ref{fig:inter_model_agreement_mtr_ib_noncausal} show that incorporating the Conditional Information Bottleneck does change the outcome in some aspects. Model agreement remains low, and the reliance on causal agents does not increase. Nevertheless, the average attribution value for the group with the highest agreement for the causal agents is significantly more negative. This decrease from about $-0.5$ to about $-3$ highlights an increased improving influence of many Causal Agents. In addition to that, the deteriorating influence of non-causal agents depicted in the bar at $r_{\text{NLL}}=\frac{0}{5}$ of Fig.~\ref{fig:inter_model_agreement_mtr_ib_noncausal} decreased from more than $4$ to less than $3.5$.
This suggests that the CIB is able to encourage the model to prioritize causally relevant information, highlighting a benefit of this approach in addressing unfaithful attribution.

\subsection{Benchmark Performance}
Tab.~\ref{table:results_performance_nuscenes} shows that the Conditional Information Bottleneck (CIB) improves QEANet across all metrics. It is worth noting that similar performance gains are observed both with and without agent information, meaning the benefit is not due to better exploitation of surrounding agent information. The performance distance $|\Delta^{m}_{\text{No-All}}|$ remains the same. This suggests that the CIB is rather improving the overall model robustness. Another indicator is the decreased performance when using only Super Agents. QEANet combined with the CIB has a smaller gap $|\Delta^{m}_{\text{Super-All}}|$ for all metrics $m$, showing reduced influence of Confounding Agents. 

For the LAformer-based architecture, the effect of the Information Bottleneck is less pronounced. Here, only the NLL can be slightly improved, while the other metrics show poor performance. The large difference between $\text{minADE}_{10}$ and $\text{minADE}_5$ performance suggests, that LAformer is overall weak at predicting the corresponding mode probabilities. 

A likely reason is the architecture, which is structurally biased toward filtering irrelevant lane segments and thus uses surrounding-agent information less.

Overall, the CIB consistently strengthens QEANet’s performance, whereas LAformer does not benefit from this extension. When it comes to leveraging surrounding agent information, these results show that the amount of improvement by a Conditional Information Bottleneck is highly dependent on the model architecture.

For the WOMD-based models, the results in Tab.~\ref{table:results_performance_waymo} show a clear picture. The incorporation of the CIB is beneficial for both architectures across all metrics. The corresponding analysis is shown in Tab.~\ref{table:results_analysis_waymo}. The gap to the super-agent performance $|\Delta^{m}_{\text{Super-All}}|$ almost vanished, indicating that the Information Bottleneck is able to improve the decision-making regarding the surrounding agent information. For both models, MTR and EDA, we can see an improvement in benchmark performance, while the performance without agents stays mostly consistent. Therefore, the gap $|\Delta^{m}_{\text{No-All}}|$ increased, indicating more selective context utilization. Furthermore, the simultaneous improvement in $\text{MissRate}$ confirms that the CIB enhances predictive accuracy without inducing mode collapse or compromising trajectory diversity.

\setlength{\tabcolsep}{4pt}
\begin{table}[h]
\centering
\begin{tabular}{ | l | c | c | c | c | c |}
 \hline
Model & $\text{minADE} \downarrow $& $\text{minFDE} \downarrow$ & $\text{MissRate} \downarrow$\\
\hline
MTR & $0.673 \pm 0.003$ & $1.377 \pm 0.009$ & $0.168 \pm 0.001$ \\
MTR + CIB & $\mathbf{0.664 \pm 0.003}$ & $\mathbf{1.354 \pm 0.004}$ & $\mathbf{0.164 \pm 0.001}$ \\
\hline
EDA & $0.654 \pm 0.004$ & $1.365 \pm 0.011$ & $0.151 \pm 0.001$ \\
EDA + CIB & $\mathbf{0.640 \pm 0.006}$ & $\mathbf{1.334 \pm 0.015}$ & $\mathbf{0.147 \pm 0.002}$ \\

 \hline
\end{tabular}
\caption{Methods trained on a 20\% subset of WOMD and evaluated on the validation split; the subsampling of Shi et al. \cite{shiMotionTransformerGlobal2023} yields a distribution close to the full dataset. The CIB consistently improves benchmark performance.}

\label{table:results_performance_waymo}
\end{table}

\setlength{\tabcolsep}{3pt}

\begin{table}[h]
\centering
\begin{tabular}{ | l | c | c | c | c | c |}
 \hline
Model & $\text{minADE} \downarrow $& $\text{minFDE} \downarrow$ & $\text{MissRate} \downarrow$\\
\hline
\rowcolor{gray!25} MTR & $0.700 \pm 0.026$ & $1.440 \pm 0.042$ & $0.164 \pm 0.003$  \\
- Super Agents &  $0.695 \pm 0.026$ & $1.436 \pm 0.040$ & $0.164 \pm 0.003$ \\
- No Agents & $0.893 \pm 0.085$ & $1.919 \pm 0.220$ & $0.248 \pm 0.059$ \\

\rowcolor{gray!25} MTR + CIB &  $\mathbf{0.683 \pm 0.009}$ & $\mathbf{1.397 \pm 0.004}$ & $\mathbf{0.160 \pm 0.004}$\\
- Super Agents & $0.683 \pm 0.007$ & $1.405 \pm 0.014$ & $0.161 \pm 0.004$ \\
- No Agents & $0.855 \pm 0.036$ & $1.819 \pm 0.106$ & $0.227 \pm 0.019$\\

\hline
\rowcolor{gray!25} EDA &  $0.673 \pm 0.003$ & $1.414 \pm 0.011$ & $0.150 \pm 0.001$\\
- Super Agents & $0.667 \pm 0.007$ & $1.397 \pm 0.017$ & $0.148 \pm 0.002$ \\
- No Agents & $0.841 \pm 0.050$ & $1.843 \pm 0.120$ & $0.220 \pm 0.030$ \\

\rowcolor{gray!25} EDA + CIB & $\mathbf{0.669 \pm 0.019}$ & $\mathbf{1.395 \pm 0.036}$ & $\mathbf{0.148 \pm 0.004}$\\
- Super Agents & $0.667 \pm 0.017$ & $1.387 \pm 0.031$ & $0.147 \pm 0.003$ \\
- No Agents & $0.840 \pm 0.044$ & $1.832 \pm 0.126$ & $0.217 \pm 0.030$ \\

 \hline
\end{tabular}
\caption{Shapley-value analysis (scenes with at most 24 agents, for compute reasons). The Information Bottleneck decreases the gap between Super Agents and the default (All-agent) performance.}
\label{table:results_analysis_waymo}
\end{table}
Extending our evaluation to the joint motion prediction task, we quantify the marginal impact of surrounding agents on coupled trajectories. During computation, target trajectories remain fixed while surrounding agents are systematically removed. As shown in Tab.~\ref{table:results_analysis_joint_waymo}, the baseline BeTop exhibits a significant performance drop when removing all non-target agents. Consistent with the marginal prediction results, utilizing only Super Agents yields the best performance, suggesting that filtering non-essential interactors reduces predictive noise. This instability in decision-making persists across all investigated models. Even joint-prediction frameworks, despite being engineered for high-interaction environments, remain unstable to these perturbations.
\begin{table}[h]
\centering

\begin{tabular}{| l | c | c | c |}
\hline

Model & $\text{minADE} \downarrow $ & $\text{minFDE} \downarrow$ & $\text{MissRate} \downarrow$\\
\hline

\rowcolor{gray!25} BeTop & $\mathbf{0.988 \pm 0.002}$ & $\mathbf{2.267 \pm 0.016}$ & $\mathbf{0.298 \pm 0.012}$ \\
- Super Agents & $0.912 \pm 0.016$ & $2.069 \pm 0.047$ & $0.272 \pm 0.017$ \\
- No Agents & $1.221 \pm 0.019$ & $2.901 \pm 0.047$ & $0.373 \pm 0.009$ \\

\rowcolor{gray!25} BeTop + CIB & $1.018 \pm 0.023$ & $2.335 \pm 0.042$ & $0.318 \pm 0.008$ \\
- Super Agents & $0.937 \pm 0.043$ & $2.143 \pm 0.113$ & $0.288 \pm 0.005$ \\
- No Agents & $1.241 \pm 0.035$ & $2.939 \pm 0.067$ & $0.383 \pm 0.004$ \\

\hline
\end{tabular}

\caption{Joint motion prediction. The marginal contribution of surrounding agents is estimated by fixing the two target trajectories and removing the remaining actors. Both variants (BeTop, BeTop+IB) were trained on the WOMD split over three seeds and evaluated on the Waymo Interaction Challenge validation split.}

\label{table:results_analysis_joint_waymo}
\end{table}

\subsection{Robustness}
\begin{table}[h]
\centering
\begin{tabular}{ | l | c | c | }
\hline
& \%Abs($\Delta$)$_{\text{Noise,0.2}}$ $\downarrow$ & \%Abs($\Delta$)$_{\text{Noise,0.4}}$ $\downarrow$ \\
\hline
QEANet & $0.75\% \pm 0.07\%$ & $1.48\% \pm 0.15\%$ \\ 
QEANet + CIB & $\mathbf{0.72\% \pm 0.05\%}$ & $\mathbf{1.33\% \pm 0.15\%}$ \\  

\hline
LAformer & $1.06\% \pm 0.06\%$ & $2.06\% \pm 0.10\%$ \\ 
LAformer + CIB & $\mathbf{0.35\% \pm 0.12\%}$ & $\mathbf{0.71\% \pm 0.24\%}$  \\  
\hline

\end{tabular}

\caption{Noise-based perturbation robustness ($\mathrm{minADE}_{10}$).}

\label{table:results_noise_pert}
\end{table}

\begin{table}[h]

\centering

\begin{tabular}{ | l | c | c | }
\hline

& \%$\text{Abs}(\Delta)_{\text{Causal}}$ $\downarrow$ & \%$\text{Abs}(\Delta)_{\text{Non-Causal}}$ $\downarrow$ \\
\hline
MTR & $61.7 \% \pm 6.3 \% $ & $12.7 \% \pm 1.7 \% $ \\ 
MTR + CIB & $\mathbf{58.1 \% \pm 1.0 \% }$  &  $\mathbf{11.6 \% \pm 0.3 \%} $ \\  
\hline

EDA & $72.7 \% \pm 1.5 \% $ &  $17.0 \% \pm 0.4 \% $  \\
EDA + CIB & $\mathbf{65.6 \% \pm 4.7 \%} $ & $\mathbf{14.8 \% \pm 0.7 \%} $ \\  
 
\hline
\end{tabular}

\caption{Removal-based perturbation on causal and non-causal agents ($\mathrm{minADE}$).}

\label{table:results_causal_attack}
\end{table}

Table~\ref{table:results_noise_pert} reports robustness under noise-based perturbations. For both QEANet and LAformer, incorporating the Conditional Information Bottleneck consistently reduces sensitivity to Gaussian noise. The effect is strongest for LAformer, where degradation is reduced by more than half compared to the baseline. 

Table~\ref{table:results_causal_attack} summarizes the removal-based perturbations. As expected, removing Causal Agents strongly degrades performance across all models. Incorporating the Conditional Information Bottleneck reduces this effect. Similarly, when removing Non-Causal Agents, models equipped with the bottleneck exhibit consistently lower sensitivity. For MTR the relative improvement is about $6\%$ (Causal) and $9\%$ (Non-Causal) while it is $10\%$ (Causal) and $13\%$ (Non-Causal) for EDA. 
As we observed an improved decision-making in Tab.~\ref{table:results_performance_waymo}, this observation can be further supported here and is in line with the reduced influence of Confounding Agents we saw in Fig.~\ref{fig:inter_model_agreement_mtr_ib_noncausal}. 

\section{Discussion}

Our investigation reveals a crucial paradox in modern trajectory prediction: while models are designed to leverage social context, many surrounding agents act as Confounders that actively degrade accuracy, often canceling out the benefits from the few truly helpful Super Agents. This inefficiency stems from an inconsistent learning process; our attribution analysis shows that models trained with different seeds learn vastly different decision-making schemes that fail to robustly align with human-annotated causal relationships. This inconsistency poses a significant reliability challenge for their deployment in safety-critical systems, as model behavior is unpredictably dependent on the specific training run.

While our proposed Conditional Information Bottleneck improves general robustness by regularizing information flow, it does not solve this core problem, as it fails to equip models with an explicit mechanism for causal reasoning or improve their ability to distinguish causal from non-causal agents. Nevertheless, it can be seen as a starting point to guide future research. Ultimately, this work's primary contribution is the exposure of this deep-seated flaw, demonstrating that simply adding more contextual information is insufficient without the ability to selectively and robustly focus on what truly matters.

\section*{Acknowledgement}
This work is part of BrainLinks-BrainTools which is funded by the Federal Ministry of Economics, Science and Arts of Baden-Württemberg within the sustainability program for projects of the excellence initiative II.
The authors acknowledge the computation time provided by bwUniCluster funded by the Ministry of Science, Research and the Arts Baden-Württemberg and the Universities of the State of Baden-Württemberg, Germany, within the framework program bwHPC.
This research was funded by the Deutsche Forschungsgemeinschaft (DFG, German Research Foundation) under grant number 539134284, through EFRE (FEIH\_2698644) and the state of Baden-Württemberg. 
\begin{center}
\raisebox{-0.5\height}{\includegraphics[width=0.42\columnwidth]{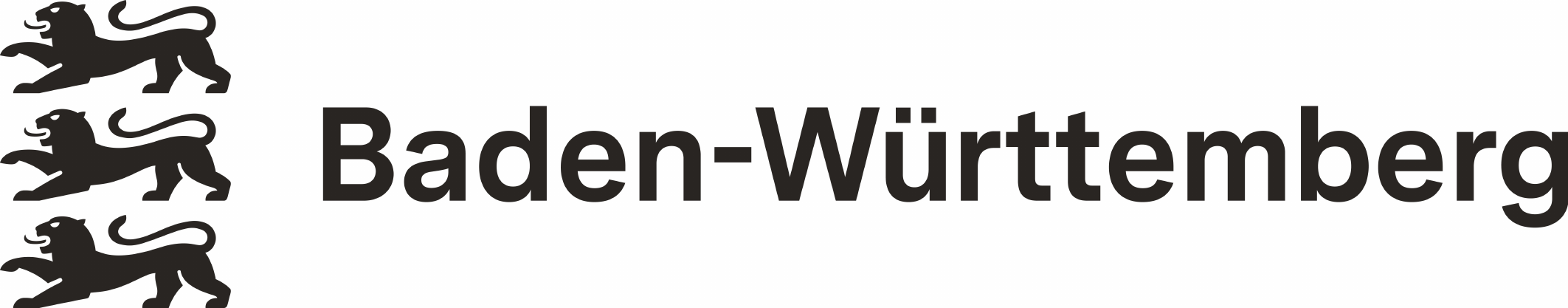}}
\hspace{0.04\columnwidth}
\raisebox{-0.5\height}{\includegraphics[width=0.42\columnwidth]{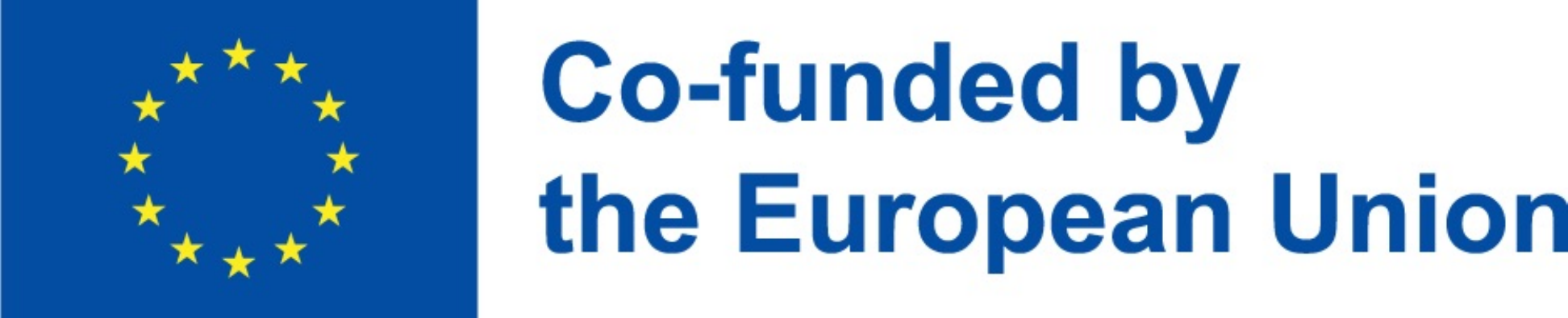}}
\end{center}

\bibliographystyle{IEEEtran}
\bibliography{bibliography}

\begin{thebibliography}{10}
\providecommand{\url}[1]{#1}
\csname url@samestyle\endcsname
\providecommand{\newblock}{\relax}
\providecommand{\bibinfo}[2]{#2}
\providecommand{\BIBentrySTDinterwordspacing}{\spaceskip=0pt\relax}
\providecommand{\BIBentryALTinterwordstretchfactor}{4}
\providecommand{\BIBentryALTinterwordspacing}{\spaceskip=\fontdimen2\font plus
\BIBentryALTinterwordstretchfactor\fontdimen3\font minus \fontdimen4\font\relax}
\providecommand{\BIBforeignlanguage}[2]{{%
\expandafter\ifx\csname l@#1\endcsname\relax
\typeout{** WARNING: IEEEtran.bst: No hyphenation pattern has been}%
\typeout{** loaded for the language `#1'. Using the pattern for}%
\typeout{** the default language instead.}%
\else
\language=\csname l@#1\endcsname
\fi
#2}}
\providecommand{\BIBdecl}{\relax}
\BIBdecl

\bibitem{chenQEANetImplicitSocial2024}
\BIBentryALTinterwordspacing
J.~Chen, Z.~Wang, J.~Wang, and B.~Cai, ``{Q-EANet}: Implicit social modeling for trajectory prediction via experience-anchored queries,'' \emph{IET Intelligent Transport Systems}, vol.~18, no.~6, pp. 1004--1015, 2024. [Online]. Available: \url{https://articlelibrary.wiley.com/doi/abs/10.1049/itr2.12477}
\BIBentrySTDinterwordspacing

\bibitem{liuLAformerTrajectoryPrediction2023}
\BIBentryALTinterwordspacing
M.~Liu, H.~Cheng, L.~Chen, H.~Broszio, J.~Li, R.~Zhao, M.~Sester, and M.~Y. Yang, ``{{LAformer}}: {{Trajectory Prediction}} for {{Autonomous Driving}} with {{Lane-Aware Scene Constraints}},'' \emph{Proceedings of the IEEE/CVF Conference on Computer Vision and Pattern Recognition (CVPR) Workshops}, 2023. [Online]. Available: \url{http://arxiv.org/abs/2302.13933}
\BIBentrySTDinterwordspacing

\bibitem{shiMTRMultiAgentMotion2024}
\BIBentryALTinterwordspacing
S.~Shi, L.~Jiang, D.~Dai, and B.~Schiele, ``{{MTR}}++: {{Multi-Agent Motion Prediction}} with {{Symmetric Scene Modeling}} and {{Guided Intention Querying}},'' \emph{IEEE Transactions on Pattern Analysis and Machine Intelligence}, 2024. [Online]. Available: \url{http://arxiv.org/abs/2306.17770}
\BIBentrySTDinterwordspacing

\bibitem{linEDAEvolvingDistinct2023}
\BIBentryALTinterwordspacing
L.~Lin, X.~Lin, T.~Lin, L.~Huang, R.~Xiong, and Y.~Wang, ``{{EDA}}: {{Evolving}} and {{Distinct Anchors}} for {{Multimodal Motion Prediction}},'' \emph{Proceedings of the AAAI Conference on Artificial Intelligence}, 2023. [Online]. Available: \url{http://arxiv.org/abs/2312.09501}
\BIBentrySTDinterwordspacing

\bibitem{roelofsCausalAgentsRobustnessBenchmark2022}
R.~Roelofs, L.~Sun, B.~Caine, K.~S. Refaat, B.~Sapp, S.~Ettinger, and W.~Chai, ``{{CausalAgents}}: {{A Robustness Benchmark}} for {{Motion Forecasting}} using {{Causal Relationships}},'' Oct. 2022.

\bibitem{zhangAdversarialRobustnessTrajectory2022}
Q.~Zhang, S.~Hu, J.~Sun, Q.~A. Chen, and Z.~M. Mao, ``On {{Adversarial Robustness}} of {{Trajectory Prediction}} for {{Autonomous Vehicles}},'' in \emph{Proceedings of the {{IEEE}}/{{CVF Conference}} on {{Computer Vision}} and {{Pattern Recognition}}}, 2022, pp. 15\,159--15\,168.

\bibitem{saadatnejadAreSociallyawareTrajectory2022}
S.~Saadatnejad, M.~Bahari, P.~Khorsandi, M.~Saneian, S.-M. {Moosavi-Dezfooli}, and A.~Alahi, ``Are socially-aware trajectory prediction models really socially-aware?'' Feb. 2022.

\bibitem{caoAdvDORealisticAdversarial2022}
Y.~Cao, C.~Xiao, A.~Anandkumar, D.~Xu, and M.~Pavone, ``{{AdvDO}}: {{Realistic Adversarial Attacks}} for {{Trajectory Prediction}},'' Sep. 2022.

\bibitem{shapley7ValueNPerson2020}
L.~Shapley, ``7. {{A Value}} for n-{{Person Games}}. {{Contributions}} to the {{Theory}} of {{Games II}} (1953) 307-317.'' in \emph{Classics in {{Game Theory}}}, H.~W. Kuhn, Ed.\hskip 1em plus 0.5em minus 0.4em\relax Princeton University Press, Nov. 2020, pp. 69--79.

\bibitem{makansiYOUMOSTLYWALK2022}
O.~Makansi, T.~Brox, and B.~Scholkopf, ``{{YOU MOSTLY WALK ALONE}}: {{ANALYZING FEATURE ATTRIBUTION IN TRAJECTORY PREDICTION}},'' in \emph{International Conference on Learning Representations (ICLR)}, 2022.

\bibitem{alemiDeepVariationalInformation2019}
\BIBentryALTinterwordspacing
A.~A. Alemi, I.~Fischer, J.~V. Dillon, and K.~Murphy, ``Deep {{Variational Information Bottleneck}},'' \emph{International Conference on Learning Representations (ICLR)}, 2019. [Online]. Available: \url{http://arxiv.org/abs/1612.00410}
\BIBentrySTDinterwordspacing

\bibitem{leeConditionalGraphInformation2023}
\BIBentryALTinterwordspacing
N.~Lee, D.~Hyun, G.~S. Na, S.~Kim, J.~Lee, and C.~Park, ``Conditional {{Graph Information Bottleneck}} for {{Molecular Relational Learning}},'' Jul. 2023. [Online]. Available: \url{http://arxiv.org/abs/2305.01520}
\BIBentrySTDinterwordspacing

\bibitem{shiMotionTransformerGlobal2023}
S.~Shi, L.~Jiang, D.~Dai, and B.~Schiele, ``Motion {Transformer} with global intention localization and local movement refinement,'' in \emph{Advances in Neural Information Processing Systems (NeurIPS)}, 2022.

\bibitem{caesarNuScenesMultimodalDataset2020}
\BIBentryALTinterwordspacing
H.~Caesar, V.~Bankiti, A.~H. Lang, S.~Vora, V.~E. Liong, Q.~Xu, A.~Krishnan, Y.~Pan, G.~Baldan, and O.~Beijbom, ``{{nuScenes}}: {{A}} multimodal dataset for autonomous driving,'' \emph{Proceedings of the IEEE/CVF Conference on Computer Vision and Pattern Recognition (CVPR)}, 2020. [Online]. Available: \url{http://arxiv.org/abs/1903.11027}
\BIBentrySTDinterwordspacing

\bibitem{ettingerLargeScaleInteractive2021}
S.~Ettinger, S.~Cheng, B.~Caine, C.~Liu, H.~Zhao, S.~Pradhan, Y.~Chai, B.~Sapp, C.~Qi, Y.~Zhou, Z.~Yang, A.~Chouard, P.~Sun, J.~Ngiam, V.~Vasudevan, A.~McCauley, J.~Shlens, and D.~Anguelov, ``Large scale interactive motion forecasting for autonomous driving: {The Waymo Open Motion Dataset},'' in \emph{Proceedings of the IEEE/CVF International Conference on Computer Vision (ICCV)}, 2021, pp. 9710--9719.

\bibitem{zhouQueryCentricTrajectoryPrediction2023}
\BIBentryALTinterwordspacing
Z.~Zhou, J.~Wang, Y.~Li, and Y.~Huang, ``Query-{{Centric Trajectory Prediction}},'' in \emph{2023 {{IEEE}}/{{CVF Conference}} on {{Computer Vision}} and {{Pattern Recognition}} ({{CVPR}})}.\hskip 1em plus 0.5em minus 0.4em\relax IEEE, 2023, pp. 17\,863--17\,873. [Online]. Available: \url{https://ieeexplore.ieee.org/document/10203873/}
\BIBentrySTDinterwordspacing

\bibitem{zhouQCNeXtNextGenerationFramework2023}
\BIBentryALTinterwordspacing
Z.~Zhou, Z.~Wen, J.~Wang, Y.-H. Li, and Y.-K. Huang, ``{{QCNeXt}}: {{A Next-Generation Framework For Joint Multi-Agent Trajectory Prediction}},'' \emph{Proceedings of the IEEE/CVF Conference on Computer Vision and Pattern Recognition (CVPR)}, 2023. [Online]. Available: \url{http://arxiv.org/abs/2306.10508}
\BIBentrySTDinterwordspacing

\bibitem{wangSimpleMultiagentJoint2024}
\BIBentryALTinterwordspacing
M.~Wang, H.~Zou, Y.~Liu, Y.~Wang, and G.~Li, ``A joint prediction method of multi-agent to reduce collision rate,'' 2024. [Online]. Available: \url{http://arxiv.org/abs/2411.07612}
\BIBentrySTDinterwordspacing

\bibitem{liuReasoningMultiAgentBehavioral}
H.~Liu, L.~Chen, Y.~Qiao, C.~Lv, and H.~Li, ``Reasoning multi-agent behavioral topology for interactive autonomous driving,'' in \emph{Proceedings of the IEEE/CVF Conference on Computer Vision and Pattern Recognition (CVPR)}, 2024, pp. 15\,918--15\,928.

\bibitem{jainAttentionNotExplanation2019}
\BIBentryALTinterwordspacing
S.~Jain and B.~C. Wallace, ``Attention is not {{Explanation}},'' \emph{Proceedings of the 2019 Conference of the North American Chapter of the Association for Computational Linguistics: Human Language Technologies, Volume 1 (Long and Short Papers)}, 2019. [Online]. Available: \url{http://arxiv.org/abs/1902.10186}
\BIBentrySTDinterwordspacing

\bibitem{adebayoSanityChecksSaliency2018}
J.~Adebayo, J.~Gilmer, M.~Muelly, I.~Goodfellow, M.~Hardt, and B.~Kim, ``Sanity {{Checks}} for {{Saliency Maps}},'' in \emph{Advances in {{Neural Information Processing Systems}}}, vol.~31.\hskip 1em plus 0.5em minus 0.4em\relax Curran Associates, Inc., 2018.

\bibitem{ahmadiCurbYourAttention2025}
\BIBentryALTinterwordspacing
E.~Ahmadi, R.~Mercurius, S.~Alizadeh, K.~Rezaee, and A.~Rasouli, ``Curb {{Your Attention}}: {{Causal Attention Gating}} for {{Robust Trajectory Prediction}} in {{Autonomous Driving}},'' \emph{Proceedings of the IEEE International Conference on Robotics and Automation (ICRA)}, 2025. [Online]. Available: \url{http://arxiv.org/abs/2410.07191}
\BIBentrySTDinterwordspacing

\bibitem{pourkeshavarzCaDeTCausalDisentanglement2024}
\BIBentryALTinterwordspacing
M.~Pourkeshavarz, J.~Zhang, and A.~Rasouli, ``{{CaDeT}}: {{A Causal Disentanglement Approach}} for {{Robust Trajectory Prediction}} in {{Autonomous Driving}},'' in \emph{2024 {{IEEE}}/{{CVF Conference}} on {{Computer Vision}} and {{Pattern Recognition}} ({{CVPR}})}.\hskip 1em plus 0.5em minus 0.4em\relax IEEE, 2024, pp. 14\,874--14\,884. [Online]. Available: \url{https://ieeexplore.ieee.org/document/10657124/}
\BIBentrySTDinterwordspacing

\bibitem{tishbyInformationBottleneckMethod2000}
\BIBentryALTinterwordspacing
N.~Tishby, F.~C. Pereira, and W.~Bialek, ``The information bottleneck method,'' 2000. [Online]. Available: \url{http://arxiv.org/abs/physics/0004057}
\BIBentrySTDinterwordspacing

\bibitem{castroPolynomialCalculationShapley2009}
J.~Castro, D.~G{\'o}mez, and J.~Tejada, ``Polynomial calculation of the {{Shapley}} value based on sampling,'' \emph{Computers \& Operations Research}, vol.~36, no.~5, pp. 1726--1730, May 2009.

\bibitem{petsiukRISERandomizedInput2018}
V.~Petsiuk, A.~Das, and K.~Saenko, ``{{RISE}}: {{Randomized Input Sampling}} for {{Explanation}} of {{Black-box Models}},'' Sep. 2018.

\bibitem{hamaDeletionInsertionTests2023}
N.~Hama, M.~Mase, and A.~B. Owen, ``Deletion and insertion tests in regression models,'' \emph{Journal of Machine Learning Research}, vol.~23, no.~1, 2022.

\bibitem{yuImprovingSubgraphRecognition2022}
\BIBentryALTinterwordspacing
J.~Yu, J.~Cao, and R.~He, ``Improving {{Subgraph Recognition}} with {{Variational Graph Information Bottleneck}},'' in \emph{2022 {{IEEE}}/{{CVF Conference}} on {{Computer Vision}} and {{Pattern Recognition}} ({{CVPR}})}.\hskip 1em plus 0.5em minus 0.4em\relax New Orleans, LA, USA: IEEE, Jun. 2022, pp. 19\,374--19\,383. [Online]. Available: \url{https://ieeexplore.ieee.org/document/9880086/}
\BIBentrySTDinterwordspacing

\bibitem{federiciLearningRobustRepresentations2020}
\BIBentryALTinterwordspacing
M.~Federici, A.~Dutta, P.~Forr{\'e}, N.~Kushman, and Z.~Akata, ``Learning {{Robust Representations}} via {{Multi-View Information Bottleneck}},'' Feb. 2020. [Online]. Available: \url{http://arxiv.org/abs/2002.07017}
\BIBentrySTDinterwordspacing

\end{thebibliography}

\end{document}